\documentclass{article}

     \usepackage[preprint]{neurips_2020}

\usepackage[utf8]{inputenc} 
\usepackage[T1]{fontenc}    
\usepackage{hyperref}       
\usepackage{url}            
\usepackage{booktabs}       
\usepackage{amsfonts}       
\usepackage{nicefrac}       
\usepackage{microtype}      
\usepackage{graphicx}
\usepackage{subfigure}
\usepackage{bm}
\usepackage{algorithm}
\usepackage{algorithmic}
\usepackage{amsmath}
\usepackage{authblk}

\title{Meta Feature Modulator for Long-tailed Recognition}

\author{Renzhen Wang$^{*}$}
\author{Kaiqin Hu\thanks{Equal contribution.}}
\author{Yanwen Zhu}
\author{Jun Shu}
\author{Qian Zhao}
\author{Deyu Meng\thanks{Corresponding author.}}
\affil{School of Mathematics and Statistics, Xi'an Jiaotong University \authorcr
  \{\tt wrzhen, richard, zywwyz\}@stu.xjtu.edu.cn \authorcr
  \tt xjtushujun@gmail.com \authorcr
  \{\tt timmy.zhaoqian, dymeng\}@mail.xjtu.edu.cn}

\begin{document}

\maketitle

\begin{abstract}
  Deep neural networks often degrade significantly when training data suffer from class imbalance problems. Existing approaches, e.g., re-sampling and re-weighting, commonly address this issue by rearranging the label distribution of training data to train the networks fitting well to the implicit balanced label distribution. However, most of them hinder the representative ability of learned features due to insufficient use of intra/inter-sample information of training data. To address this issue, we propose meta feature modulator (MFM), a meta-learning framework to model the difference between the long-tailed training data and the balanced meta data from the perspective of representation learning. Concretely, we employ learnable hyper-parameters (dubbed modulation parameters) to adaptively scale and shift the intermediate features of classification networks, and the modulation parameters are optimized together with the classification network parameters guided by a small amount of balanced meta data. We further design a modulator network to guide the generation of the modulation parameters, and such a meta-learner can be readily adapted to train the classification network on other long-tailed datasets. Extensive experiments on benchmark vision datasets substantiate the superiority of our approach on long-tailed recognition tasks beyond other state-of-the-art methods.

\end{abstract}
\section{Introduction}
Large-scale datasets \cite{deng2009imagenet, lin2014microsoft, zhou2017places} play an important role in visual recognition research, especially for deep learning \cite{krizhevsky2012imagenet, he2016deep}. Such datasets usually exhibit roughly uniform distributions of class labels on both training and test data, visual phenomena, however, always follow skewed distributions in real-world, which leads to inconsistent distributions between the collected training and test sets  \cite{quattoni2009recognizing, van2018inaturalist}. This dataset bias problem commonly makes deep networks easily over-fit to the head classes yet under-fit to the tail classes, due to their powerful capacity capturing dataset bias in internal representations \cite{torralba2011unbiased}. In Fig. \ref{fig_flowchart} (a), we illustrate this phenomenon of ResNet-32 \cite{he2016deep} on long-tailed CIFAR-10 \cite{cui2019class}.

To address this challenge, recent studies have mainly pursued along the line of class re-balancing, including re-sampling and re-weighting \cite{buda2018systematic, johnson2019survey}. Re-sampling strategies balance the training data distribution by over-sampling samples of tail classes \cite{chawla2002smote, han2005borderline} or under-sampling ones of head classes \cite{drummond2003c4, he2009learning}, and re-weighting strategies address the task by assigning weights for different classes \cite{mikolov2013distributed, huang2016learning, Khan2017Cost, cui2019class, cao2019learning} or different instances \cite{lin2017focal, li2019gradient, Khan2019Striking, ren2018learning, shu2019meta}. Despite sound performance, Zhou et al. \cite{zhou2019bbn} pointed out that re-balancing significantly promotes the discriminative capability, which, however, simultaneously damages the representative capability of deep networks. On this count, they proposed a bilateral-branch network (BBN) that differs from previous representation learning based methods \cite{huang2016learning, zhang2017range, cao2019learning} addressing class imbalance problem by designing specific losses, where the proposed BBN simultaneously learns the representation and discrimination of deep networks by a bilateral-branch architecture. Similarly, Kang et al. \cite{kang2020decoupling} recently proposed to decouple the training phase into representation and classifier learning, resulting in a significant improvement of long-tailed recognition. Both methods reveal that learning from the feature space and label space can efficiently improve the long-tailed recognition accuracy, and a reasonable combination of the two strategies could be mutually beneficial for deep networks to learn the representation and discrimination.

\begin{minipage}{0.54\linewidth}
Motivated by this, this paper proposes meta feature modulator (MFM), a meta-learning framework to adaptively modulate the intermediate features of deep networks for long-tailed recognition tasks. Concretely, we introduce learnable modulation parameters, acting as meta modulator, to channel-wisely scale or shift the intermediate features of classification network during training. Guided by a small amount of balanced meta-dataset, the modulator aims to train the classification network on long-tailed training data for adapting it to fit well on the implicit class-balanced label distribution (via empirical risk minimization of meta data, of which the features are not modulated). As shown in Fig. \ref{fig_flowchart}(b), the classifier ResNet-32 \cite{he2016deep} trained with the proposed MFM on an extremely long-tailed CIFAR-10 set, i.e., with an imbalance factor of 100 \cite{cui2019class}, basically maintains the label distribution of test data (a truncated heavy-tailed distribution). However, the one directly trained (Fig. \ref{fig_flowchart}(a)) mistakenly follows a long-tailed distribution. It reveals that our approach tends to approximately equal preference to all ten classes. In particular, our method needs not conditional information and feature modulation during inference, which differs from the other feature manipulation approaches, including conditional batch normalization \cite{dumoulin2016learned, ioffe2015batch} and channel attention strategies \cite{chen2017sca, hu2018squeeze}. Instead of designing specific architectures \cite{zhou2019bbn} or decoupling parameter training \cite{kang2020decoupling}, our MFM can be equipped with any existing networks, and their parameters together with the learnable modulation parameters are optimized in a unified meta-learning framework.
\end{minipage}\ \ \ \
\begin{minipage}{0.42\textwidth}
\begin{figure}[H]
  \centering
  \vspace{-4mm}
  \includegraphics[width=0.96\linewidth]{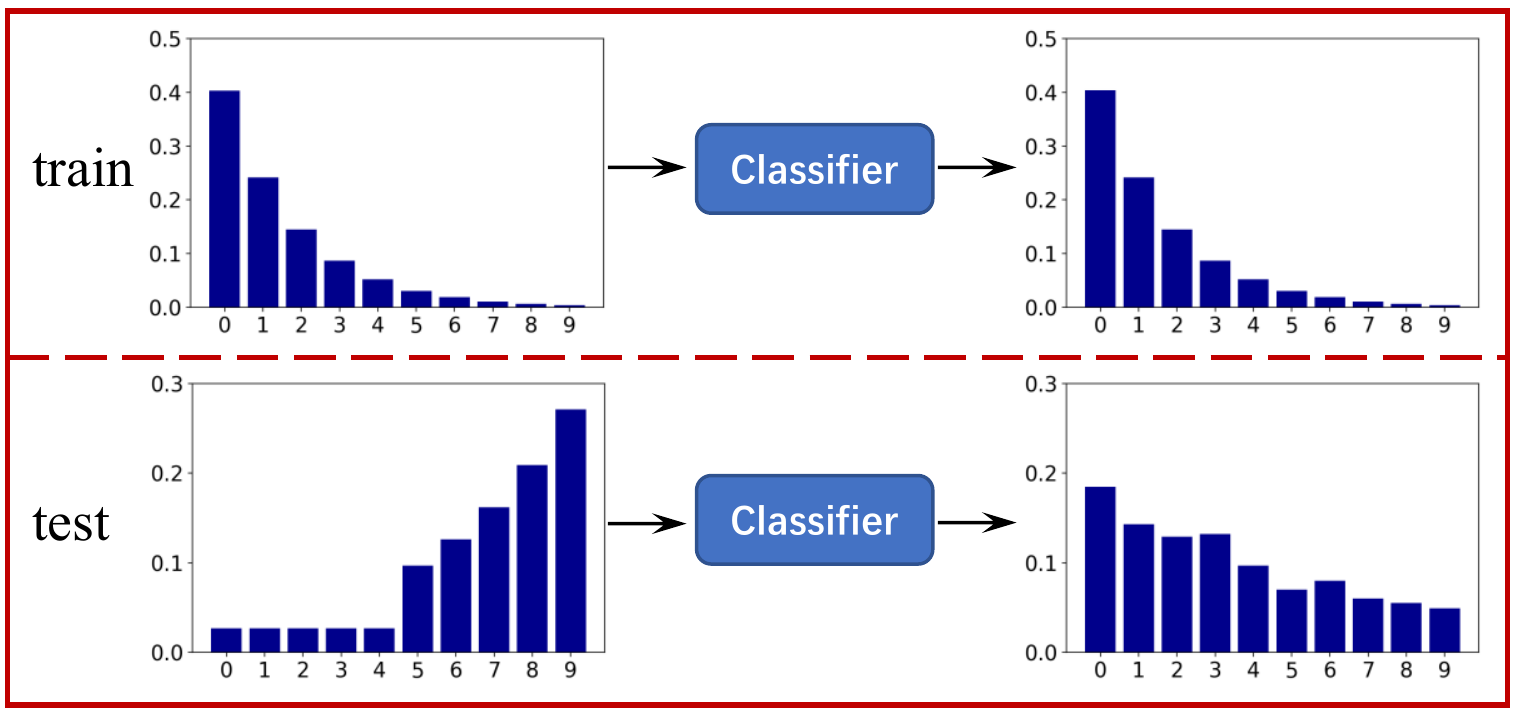}\\{(a)}
  \centering
  \includegraphics[width=0.96\linewidth]{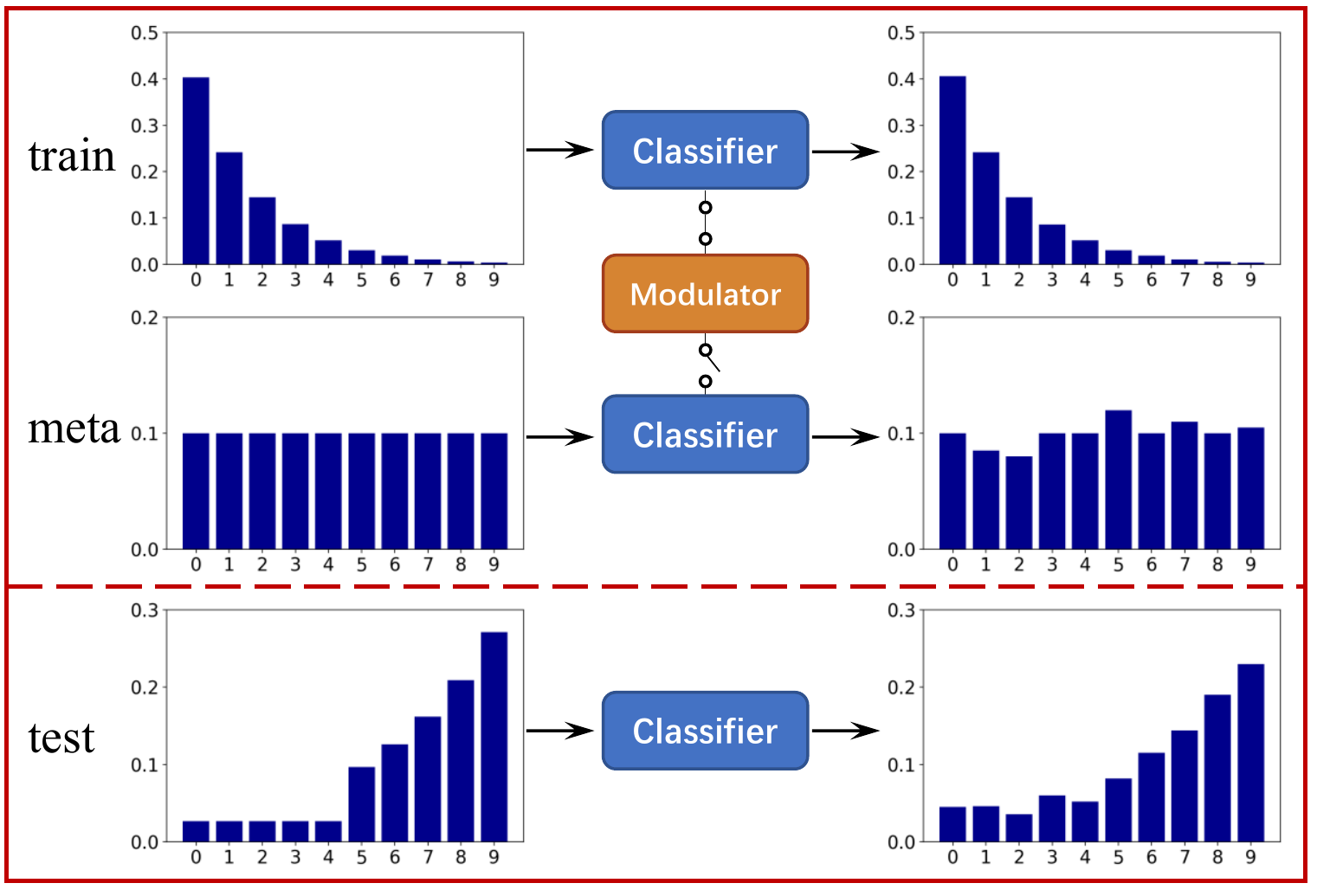}\\ {(b)}
  \vspace{-2mm}
  \label{fig_flowchart}
  \caption{The flowcharts of (a) the standard ResNet-32 and (b) MFM, where the state of modulator controlled by a switch function for feature transferability. The histograms display the label distributions, i.e, X-axis and Y-axis denote class index and per-class sample ratio (left: original, right: predicted), respectively. Both models are trained on the same long-tailed CIFAR-10 \cite{cui2019class}, and tested on a specific set with a truncated heavy-tailed label distribution.}
\end{figure}
\end{minipage}

Furthermore, we parameterize the meta modulator as an explicit network mapping, referred to as modulator network, to directly learn the scaling and shifting parameters for ameliorating the phase of representation learning. We design and conduct extensive experiments on different class-imbalanced datasets, verifying the superiority of our approach for long-tailed visual recognition. The learned modulation parameters present obvious structural information w.r.t class label, indicating its intrinsic function as a modulator for balancing the contribution per class during the training phase. Beyond that, we transfer the modulator network learned from one dataset to modulate the features of classification network during training in another long-tailed dataset, which as well significantly improves the recognition performance, showing good generalization capability of the proposed MFM.

In summary, our contributions are mainly three-fold: 1) We propose to adaptively modulate the features of classification networks during the training phase for addressing long-tailed visual recognition using a meta-learning framework. 2) Our proposed MFM is architecture-agnostic, which can be easily implemented on any off-the-shelf deep network. 3) We further design a modulator network to guide the generation of the modulation parameters, such a pre-learned meta-learner can be readily adapted to train the classification network on other long-tailed datasets.
\section{Related Work}
\textbf{Class imbalance:} Recent methods for class imbalance have mainly raised along three lines: class re-balancing, representation learning and transfer learning. Class re-balancing strategies include two categories: 1) sampling methods, which aim to balance the data distribution by over-sampling for the minority classes \cite{chawla2002smote, han2005borderline}, or under-sampling for the majority classes \cite{drummond2003c4, he2009learning}. This implies that the models trained by duplicating tailed samples might lead to over-fitting upon minority classes, or by discarding headed samples could under-represent the majority classes \cite{chawla2002smote, cui2019class}. 2) re-weighting methods aim to assign weights for different classes \cite{mikolov2013distributed, huang2016learning, Khan2017Cost, cui2019class, cao2019learning} or even different instances \cite{lin2017focal, li2019gradient, Khan2019Striking, ren2018learning, shu2019meta}, where the weighting function is usually predefined based on data distribution or adaptively learned based on the training loss. These re-balancing methods, however, damage the representative capability  when promoting the discriminative ability of deep networks \cite{zhou2019bbn}.

Representation learning based methods usually pursue the line of learning specific metrics for maintaining intra-class clusters and inter-class margins, e.g., triple-header loss \cite{huang2016learning}, range loss\cite{zhang2017range} and label-distribution-aware margin loss \cite{cao2019learning}. Recently, Zhou et al. \cite{zhou2019bbn} proposed BBN, a bilateral-branch network to simultaneously focus on learning the representation and discrimination. Kang et al. \cite{kang2020decoupling} decoupled long-tail recognition into learning representations and classification. Instead, the proposed MFM modulates the deep features of classification network by injecting meta information during training, which requires not to carefully design losses or architectures of classification network.

Transfer learning involved in long-tailed tasks aims to transfer knowledge from the head to the tail classes. The notable studies include transferring model parameters \cite{wang2017learning}, transferring the intra-class variance \cite{yin2019feature} and transferring semantic deep features \cite{liu2019large}. Our proposed MFM differs from these methods by explicitly learning a modulator mapping to balance the contribution of each class during training, and the learned modulator can be adapted to train other long-tailed datasets.

\textbf{Feature modulation:}
Our study is heavily inspired by conditional normalization \cite{dumoulin2016learned} where the affine transformation parameters in original batch normalization (BN) \cite{ioffe2015batch} are learned by some conditional information, and its generalized version FiLM \cite{perez2018film} directly conducting feature-wise affine transformation on intermediate features of a network. Along this line, various remarkable works have been proposed, including Conditional Instance Norm \cite{dumoulin2016learned, ghiasi2017exploring}, Adaptive Instance Norm \cite{huang2017arbitrary, karras2019style}, Dynamic Layer Norm\cite{kim2017dynamic}, Conditional Batch Norm \cite{de2017modulating}, Spatially Adaptive Denormalization \cite{park2019semantic} and Positional Normalization \cite{li2019positional}. Beyond that, channel attention based methods \cite{chen2017sca, hu2018squeeze} amount to another line of methods for modulating the network features, where the intermediate features are channel-wisely scaled by the conditional information. The main difference between these methods above and our MFM is that we just modulate the features during training, since our objective is to train the classification network on long-tailed training data, such that adapting the model parameters to finely fit the implicit balanced data distribution. From a meta-learning perspective, multimodal model-agnostic meta-learning (MMAML) \cite{vuorio2019multimodal} introduced task-specific parameters to adjust the task network, and context adaptation via meta-learning (CAVIA) \cite{zintgraf2019fast} proposed to modulate the features by introduced context parameters. Both methods are based on MAML \cite{finn2017model} for few-shot learning and modulate the task network from parameter space. The proposed MFM, however, seeks to directly modulate the features of classification network during the training phase, and the network is not conditioned on task-specific information during inference.

\section{Methodology}
\subsection{Meta feature modulator}
In visual recognition problems, the goal is to seek a classifier according to the given training set $\{x_i, y_i\}_{i=1}^N$, where $x_i$ is the $i$-th sample and $y_i$ is the associated label vector over $c$ classes. From the perspective of deep learning, the classification network $f(x;\bm w)$ parameterized with $\bm w$ is usually optimized by empirical risk minimization over the training set to seek the optimal parameters $\bm w^*$, i.e., $\bm {w}^*=\arg\min_{\bm w}\frac{1}{N}\sum_{i=1}^N\ell(y_i,f(x_i;\bm w))$. For notation convenience, we reformulate the deep neural network $f$ as $f(x)=f_k(g_k(x;\bm w^{(1)});\bm w^{(2)})$, where $g_k$ parameterized with $\bm w^{(1)}$ denotes the part before the $k$-th layer of the network, and $f_k$ parameterized with $\bm w^{(2)}$ denotes the rest part of $f$.

\begin{figure}[t]
	\centering
	\vspace{-2mm}
	\includegraphics[width=0.85\linewidth]{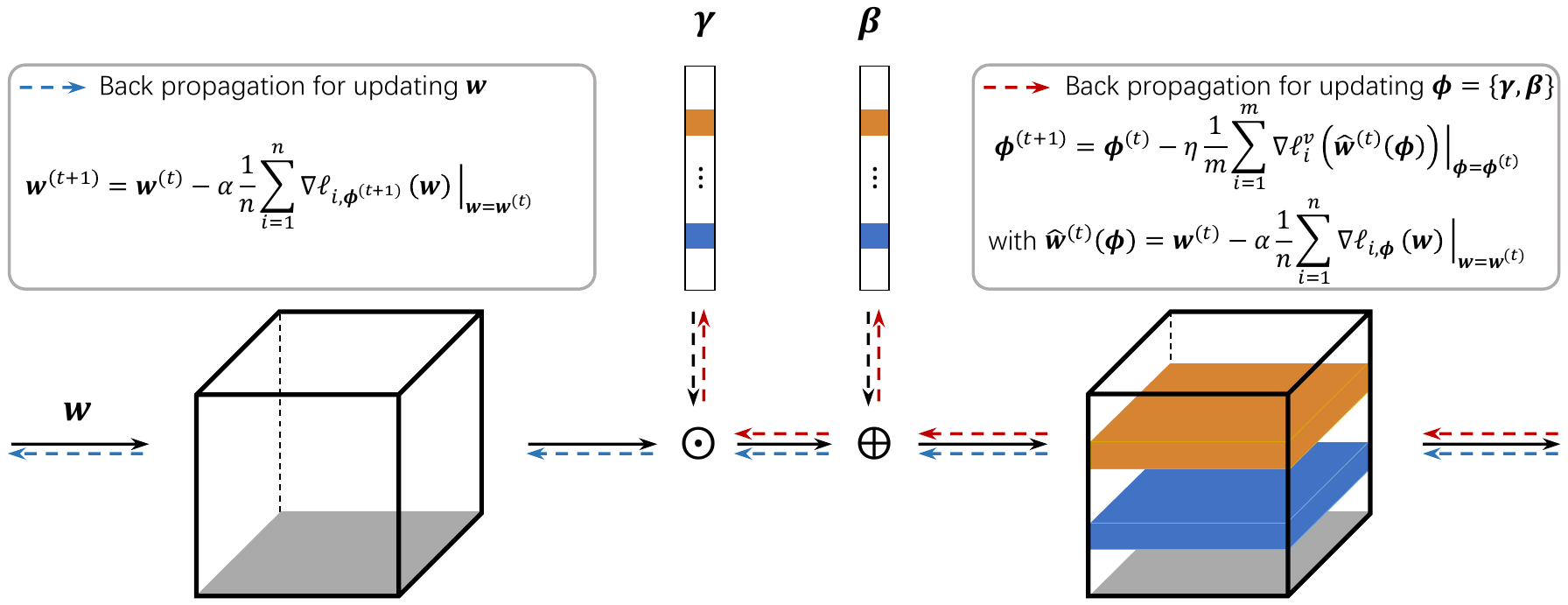}
	\vspace{-2mm}
	\caption{Visualization of training the meta feature modulator. During training, features of training data propagated along the classification network (black solid line) are channel-wisely scaled and shifted (black dotted line) by the learnable parameters $\bm\phi=\{\bm\gamma,\bm\beta\}$ for minimizing the training empirical risk Eq. \ref{eq_train}, whereas meta data only propagate along the black solid line for minimizing the meta empirical risk Eq. \ref{eq_meta}. The modulation parameters $\bm\phi$ and classification network parameters $\bm w$ are updated by gradient descent along the red and blue dotted lines, respectively. Once the model are trained, the feature modulator will not be required during inference.}
\vspace{-4mm}
	\label{fig_framework}
\end{figure}

As head classes dominate the training procedure, traditional classification networks tend to perform well on these classes, whereas degrade significantly for tail classes. To address this issue, we propose a meta modulator that directly modulates the training phase guided by a meta set (a small set of data split or copied from training data with equal samples per class), such that the classification network tends to assimilate preference for all classes. Specifically, such a modulator is implemented by scaling and shifting the intermediate features of classification network via learnable parameters. Without loss of generality, we only consider modulating features of $k$-th layer. We then have
\begin{equation}
\vspace{-1mm}
f(x)=f_k(\bm\gamma_i\odot g_k(x_i;\bm w^{(1)})\oplus\bm\beta_i;\bm w^{(2)}),
\vspace{-1mm}
\end{equation}
where $\odot$ and $\oplus$ are scaling and shifting operators, which channel-wisely (or neuron-wisely for a fully-connected layer feature) modulate the features, as shown in Fig. \ref{fig_framework}. For simplicity, we abbreviate the formulation above as $f(x)=f_k(\bm\phi_i\circ g_k(x_i;\bm w^{(1)});\bm w^{(2)})$ with $\bm\phi_i=\{\bm\gamma_i,\bm\beta_i\}$.
Formally, we minimize the following empirical risk problem over training set:
\begin{equation}
\bm w^*(\bm \phi)=\arg\min_{\bm w}\frac{1}{N}\sum_{i=1}^N\ell(y_i,f_k(\bm\phi_i\circ g_k(x_i;\bm w^{(1)});\bm w^{(2)})).
\label{eq_train}
\end{equation}
Note that the weight vector sequences $\bm\phi=\{\bm\phi_i\}_{i=1}^N$ are unknown upon beginning and treated as learnable hyper-parameters in our method. We aim to automatically learn them by a meta-learning manner where the optimal selection of $\bm\phi$ is based on empirical risk minimization of the meta data $\{x_i^v,y_i^v\}_{i=1}^M$. Benefiting from such a class-balanced meta set, we thus expect that the model is guided to tend to equal preference for each class. This can be formally formulated as
\begin{small}
\begin{equation}
\bm\phi^*=\arg\min_{\bm\phi}\frac{1}{M}\sum_{i=1}^M \ell(y_i^v,f_k( g_k(x_i^v;{\bm w^*}^{(1)}(\bm\phi));{\bm w^*}^{(2)}(\bm\phi))).
\label{eq_meta}
\end{equation}
\end{small}
Optimizing Eq. (\ref{eq_train})-(\ref{eq_meta}) involves a bilevel programming and could be directly solved using the recent meta-learning strategies \cite{andrychowicz2016learning, franceschi2018bilevel, shu2019meta}. We illustrate the optimization procedure in Fig. \ref{fig_framework}.

\subsection{Modulator network}
Instead of treating the modulation parameters as trainable variables, we further introduce another network, i.e., modulator network, to generate the modulation parameters $\bm\phi$, which is inspired by conditional batch normalization \cite{dumoulin2016learned, ioffe2015batch}. The modulator network can conveniently extend the modulation operation to more strategies, like FiLM \cite{perez2018film} and attention-based modulation \cite{mnih2014recurrent, vaswani2017attention}. More importantly, such a pre-trained modulator network can act as a meta-learner being adapted to train the classification network across datasets (in Section 4.3).

Here we adopt a single multilayer perceptron (MLP) with one hidden layer to output both $\bm\gamma_i$ and $\bm\beta_i$ for $i$-th training sample. For multi-layer feature modulation, we output a single vector per sample and dimidiate it as $\bm\gamma_i$ and $\bm\beta_i$, and then reshape them for different layers. When the classification network is wide, the modulator network needs to output large amount of modulation parameters. For example, 4096*2 modulation parameters are required for modulating the features output by 4096-neuron fully-connected (FC) layer of VGG \cite{simonyan2014very}. This implies a large-scale modulator network is required, which may arise over-fitting. Inspired by previous works \cite{chen2015compressing, shaban2017one}, we introduce a weight hashing layer (WH-layer), which is parameter-free and could map a low-dimensional vector output by the original modulator network to high-dimensional modulation parameters $\{\bm\gamma_i,\bm\beta_i\}$. Please refer to supplementary materials for more details about the weight hashing and the modulator network.

The next key is how to set the input of this modulator network. For the $i$-th sample, our modulator network takes the soft label $\hat{y}_i$ directly produced by the classification network (without feature modulation) as the input. We denote the modulator network as $\mathcal M$, and then we have $\{\bm\gamma_i,\bm\beta_i\}=\mathcal M(\hat y_i;\bm\phi)$, where $\bm\phi$ is the parameters of $\mathcal M$. As a result, the modulated classification network (abbreviated as MCN to distinguish from the classification network) has the following formulation:
\begin{equation}
f(x)=f_k(\mathcal M(\hat y_i;\bm\phi)\circ g_k(x_i;\bm w^{(1)});\bm w^{(2)}).
\label{eq_modulator}
\end{equation}
Here we present an intuitive explanation about the formulation:  The classification network aims to model the target balanced label distribution from Eq. (\ref{eq_meta}), and the MCN expects to fit the long-tailed label distribution of training data from Eq. (\ref{eq_train}). Since the two distributions are not consistent and the soft label produced by classification network is probably misled, the MCN seeks to correct it conditioned on the sample itself and its soft label, which complies with the reality that $p(y|x)=\int_{\hat y} p(y|x,\hat y)p(\hat y|x)\mathrm d\hat y$, where $p(\hat y|x)$ corresponds to the feed-forward procedure for predicting soft labels. As for optimization of $\bm\phi$ and $\bm w$, we substitute $\bm\phi_i=\mathcal M(\hat y_i;\bm\phi)$ to Eq. (\ref{eq_train})-(\ref{eq_meta}), and the resulting bilevel problem can be solved similar to \cite{andrychowicz2016learning, franceschi2018bilevel, shu2019meta}.

\subsection{Meta-learning optimization}
\begin{algorithm}[t]
   \caption{Meta feature modulator training algorithm}
   \label{algorithm}
   {\bfseries Input:} Training data $\mathcal D$, meta data $\mathcal{V}$, batch size $n,m$, max iterations $T$\\
   {\bfseries Output:} Parameters $\bm w^{(T)}$ of classification network $f$
\begin{algorithmic}[1]
   \STATE Initialize classification network parameters $\bm w^{(0)}$, and modulator network parameters $\bm\phi^{(0)}$.
   \FOR{$t=1$ {\bfseries to} $T-1$}
   \STATE $\mathcal D_n=\{x_i,y_i\}_{i=1}^n \leftarrow \mbox{SampleMiniBatch}(\mathcal D, n)$.
   \STATE $\mathcal V_m=\{x_i^v,y_i^v\}_{i=1}^m \leftarrow \mbox{SampleMiniBatch}(\mathcal V, m)$.
   \STATE Formulate the classification network learning function $\hat{\bm w}^{(t)}$ by Eq. \ref{eq_bp}.
   \STATE Update $\bm\phi^{(t+1)}$ by Eq. \ref{eq_modulator_bp} on $\mathcal V_m$.
   \STATE Update $\bm w^{(t+1)}$ by Eq. \ref{eq_classifier_bp} on $\mathcal D_n$.
   \ENDFOR
\end{algorithmic}
\end{algorithm}
Optimizing the parameters of classification network and modulator network often involves two nested loops. Here we employ an online one-loop optimization strategy and take stochastic gradient descent (SGD) as an example optimizer. At every training step $t$, we optimize the training loss Eq. \ref{eq_train} on a mini-batch training set $\{(x_i, y_i)\}_{i=1}^n$, where $n$ is the mini-batch size. The resulting parameters of classification network are w.r.t that of the modulator network, i.e.,
\begin{equation}
\label{eq_bp} \hat{\bm w}^{(t)}(\bm\phi)=\bm w^{(t)}-\alpha\frac{1}{n}\sum_{i=1}^n\nabla\ell_{i,\bm\phi}(\bm w)|_{\bm w=\bm w^{(t)}},
\end{equation}
where $\ell_{i,\bm\phi}(\bm w)=\ell(y_i,f_k(\mathcal M(\hat y_i^{(t)};\bm\phi)\circ g_k(x_i;{\bm w^{(1)}});{\bm w^{(2)}}))$ and $\alpha$ is the descent step size. Then we update the parameter $\bm\phi$ of modulator network by moving it along the objective gradient of Eq. \ref{eq_bp} on a mini-batch meta set $\{x_i^v,y_i^v\}_{i=1}^m$:
\begin{equation}
\label{eq_modulator_bp}
\bm\phi^{(t+1)}=\bm\phi^{(t)}-\eta\frac{1}{m}\sum_{i=1}^m\nabla\ell_i^v(\hat{\bm w}^{(t)}(\bm\phi))|_{\bm\phi=\bm\phi^{(t)}},
\end{equation}
where $\ell_i^v(\hat{\bm w}^{(t)}(\bm\phi))=\ell(y_i^v,f_k( g_k(x_i^v;{\hat{\bm w}^{(t)^{(1)}}}(\bm\phi));{\hat{\bm w}^{(t)^{(2)}}}(\bm\phi)))$ and $\eta$ is the descent step size of $\bm\phi$. With the updated $\bm\phi^{(t+1)}$, the parameter $\bm w$ of the classification network can be ameliorated as
\begin{equation}
\label{eq_classifier_bp}
\bm w^{(t+1)}=\bm w^{(t)}-\alpha\frac{1}{n}\sum_{i=1}^n\nabla\ell_{i,\bm\phi^{(t+1)}}(\bm w)|_{\bm w=\bm w^{(t)}},
\end{equation}
where $\bm\phi^{(t+1)}$ is a quantity rather than learnable variable. The pseudo code is listed in Algorithm \ref{algorithm}.

\section{Experimental Evaluations}
\subsection{Experimental results on long-tailed CIFAR}
We start our evaluation on the benchmark datasets CIFAR-10 and CIFAR-100 \cite{krizhevsky2009learning}, which contain 50,000/10,000 training/validation images of size $32\times32$ with 10 and 100 classes, respectively. Following the setup in \cite{cui2019class}, we experiment with Long-Tailed CIFAR dataset, where the imbalance factor is defined as the ratio between the most frequent class and the least frequent class, and the number of training samples follows an exponential decay across different classes. As for the meta data, we adopt two strategies: 1) randomly sampling 10 samples per class from original training set followed as \cite{shu2019meta}, which do not intersect with the data used to train the network; 2) randomly duplicating 20 samples (or the number of the least frequent class when it is less than 20) per class in the training set, which are shared with the current training set and referred to as as development data to avoid confusion. We use PyTorch platform \cite{paszke2017automatic}, and just modulate the last convolution layer if not specifically mentioned (see appendix for more training details). We employ ResNet-32 \cite{he2016deep} with cross-entropy loss as our baseline. Comparison methods include: 1) \textbf{BaseModel}, the baseline model directly trained on the training set; 2) \textbf{Focal} \cite{lin2017focal}, \textbf{CB}\cite{cui2019class} and \textbf{LDAM-DRW} \cite{cao2019learning}, three state-of-the-art sample or class re-weighting methods; 3) \textbf{BBN} \cite{zhou2019bbn}, representation learning method for improving the long-tailed recognition; 4) \textbf{L2RW} \cite{ren2018learning} and \textbf{MW-Net} \cite{shu2019meta}, two re-weighting methods based on meta-learning, where an additional meta set is required to assign weights for training samples.

\begin{figure}[t]
  \begin{minipage}{0.32\linewidth}
  \centering
  \includegraphics[width=0.9\linewidth]{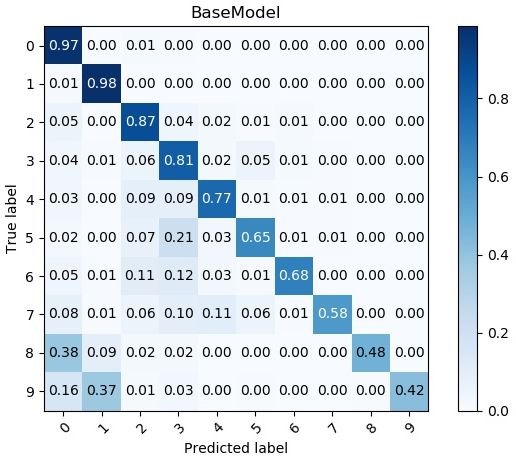}\\
  \end{minipage}
  \begin{minipage}{0.33\linewidth}
  \centering
  \includegraphics[width=0.9\linewidth]{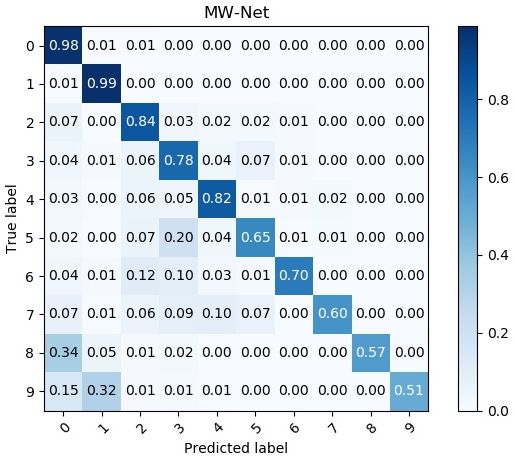}\\
  \end{minipage}
  \begin{minipage}{0.32\linewidth}
  \centering
  \includegraphics[width=0.9\linewidth]{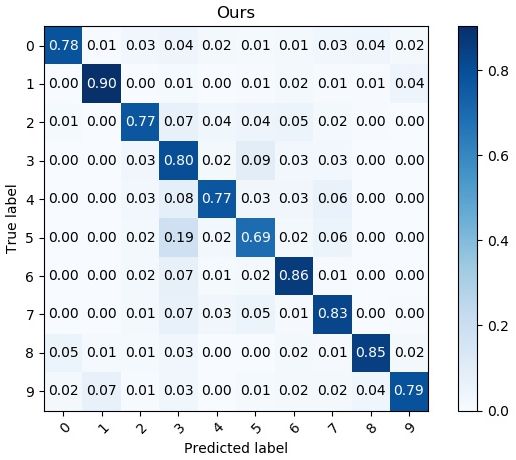}\\
  \end{minipage}\vspace{-2mm}
  \caption{Confusion matrices of BaseModel, MW-Net \cite{shu2019meta}, and ours on long-tailed CIFAR-10.}
  \label{fig_confusion_matrix}
\end{figure}

\begin{table}[t]
\vspace{-1mm}
\caption{Top-1 error rates of ResNet-32 on long-tailed CIFAR-10 and CIFAR-100.}\vspace{-2mm}
  \label{table_imbalance}
  \centering
  \begin{tabular}{c|c|c|c|c|c|c}
  \toprule
  Dataset Name & \multicolumn{3}{c|}{Long-Tailed CIFAR-10} & \multicolumn{3}{c}{Long-Tailed CIFAR-100}\\
  \hline
  Imbalance & 100 & 50 & 10 & 100 & 50 & 10 \\
  \hline
  BaseModel & 26.90& 22.31& 13.55& 63.78& 56.65 &43.15\\
  Focal \cite{lin2017focal}& 29.62& 23.28& 13.34& 61.59& 55.68& 44.22\\
  CB \cite{cui2019class} & 25.43& 20.73& 12.90& 60.40& 54.83& 42.01\\
  LDAM-DRW \cite{cao2019learning} & 22.97& 18.97& 11.84& 57.96& 53.38& 41.29\\
  BBN \cite{zhou2019bbn} &20.18 &17.82 &11.68 &57.44 &52.98 &40.88\\
  \hline
  \hline
  L2RW \cite{ren2018learning}& 25.84& 21.07& 17.88& 59.77& 55.56& 46.27\\
  MW-Net \cite{shu2019meta}& 24.79& 19.94& 12.16& 57.91& 53.26& 41.54\\
  Ours (meta)& 19.83& 18.09& 11.21& \textbf{56.09}& \textbf{50.50}& 40.26\\
  Ours (development)& \textbf{19.72}& \textbf{16.16}& \textbf{11.07}& 56.40& 51.17& \textbf{39.48}\\
  \bottomrule
  \end{tabular}
  \vspace{-4mm}
\end{table}

Table \ref{table_imbalance} reports the top-1 error rates of various methods on long-tailed CIFAR-10 and CIFAR-100 under three different imbalance factors: 10 ,50 and 100. As shown, our MFM consistently attains the best performance across all the datasets. Especially compared with the meta-learning-based methods L2RW\cite{ren2018learning} and MW-Net \cite{shu2019meta}, we achieved a significant improvement under all the experimental settings. In an extreme imbalance scenario, i.e., long-tailed CIFAR-10 with an imbalance factor of 100, we achieve 19.72\% error rate, which is over 5.07\% lower than that of methods based on meta-learning \cite{ren2018learning,shu2019meta}. In Fig. \ref{fig_confusion_matrix}, we visualize the confusion matrices for BaseModel, MW-Net and the proposed MFM, which shows that our MFM has a very significant improvement on recall for the least frequent class, and even tends to basically equal preference for all classes. In Fig. \ref{fig_recall}, we further compare per-class recall of trained networks on an imbalanced test set with truncated heavy-tailed label distribution (the same as Fig. \ref{fig_flowchart}, please refer to supplementary materials for detailed experimental settings and more results), and the results show superior generalization of our MFM.

\subsection{What do modulator network learn?}
\begin{figure}[t]
\begin{minipage}{0.44\linewidth}\vspace{-2mm}
  \centering
    \includegraphics[width=0.9\linewidth]{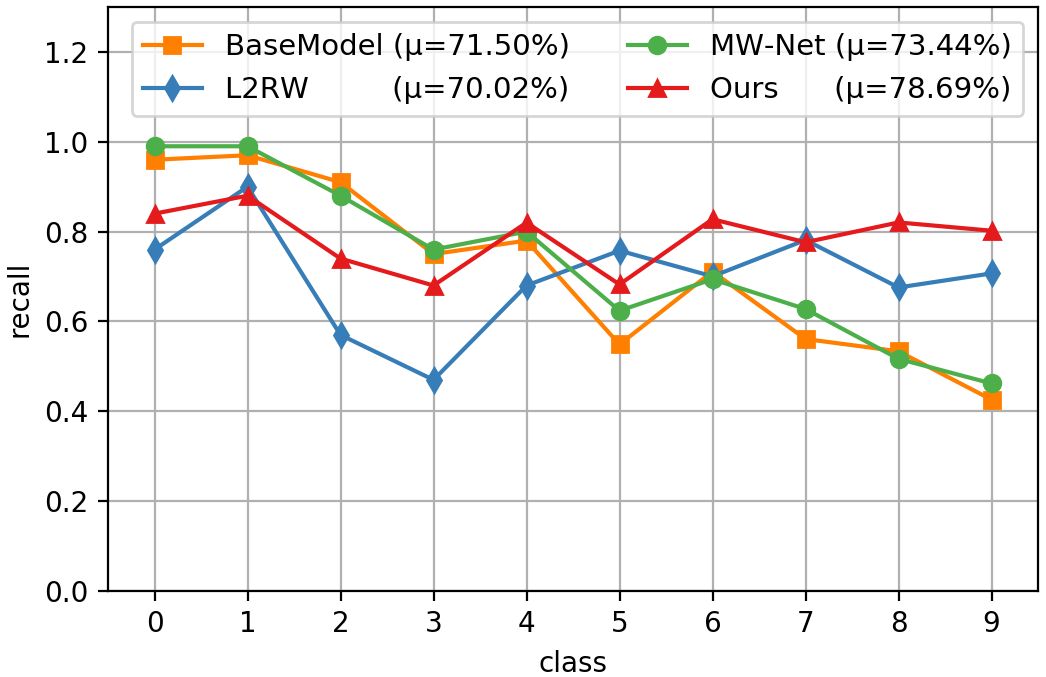}
   \vspace{-3mm}
  \caption{Per-class recall comparisons on imbalanced test dataset, the classifier is trained on long-tailed CIFAR-10 with an imbalanced factor of 100, where $\mu$ denotes mean recall.} \label{fig_recall}
  \vspace{-1mm}
    \end{minipage}\vspace{0mm}\ \ \ \ \ \
\begin{minipage}{0.53\linewidth}\vspace{-2mm}
\begin{minipage}{0.47\linewidth}
     \centering
    \includegraphics[width=1\linewidth]{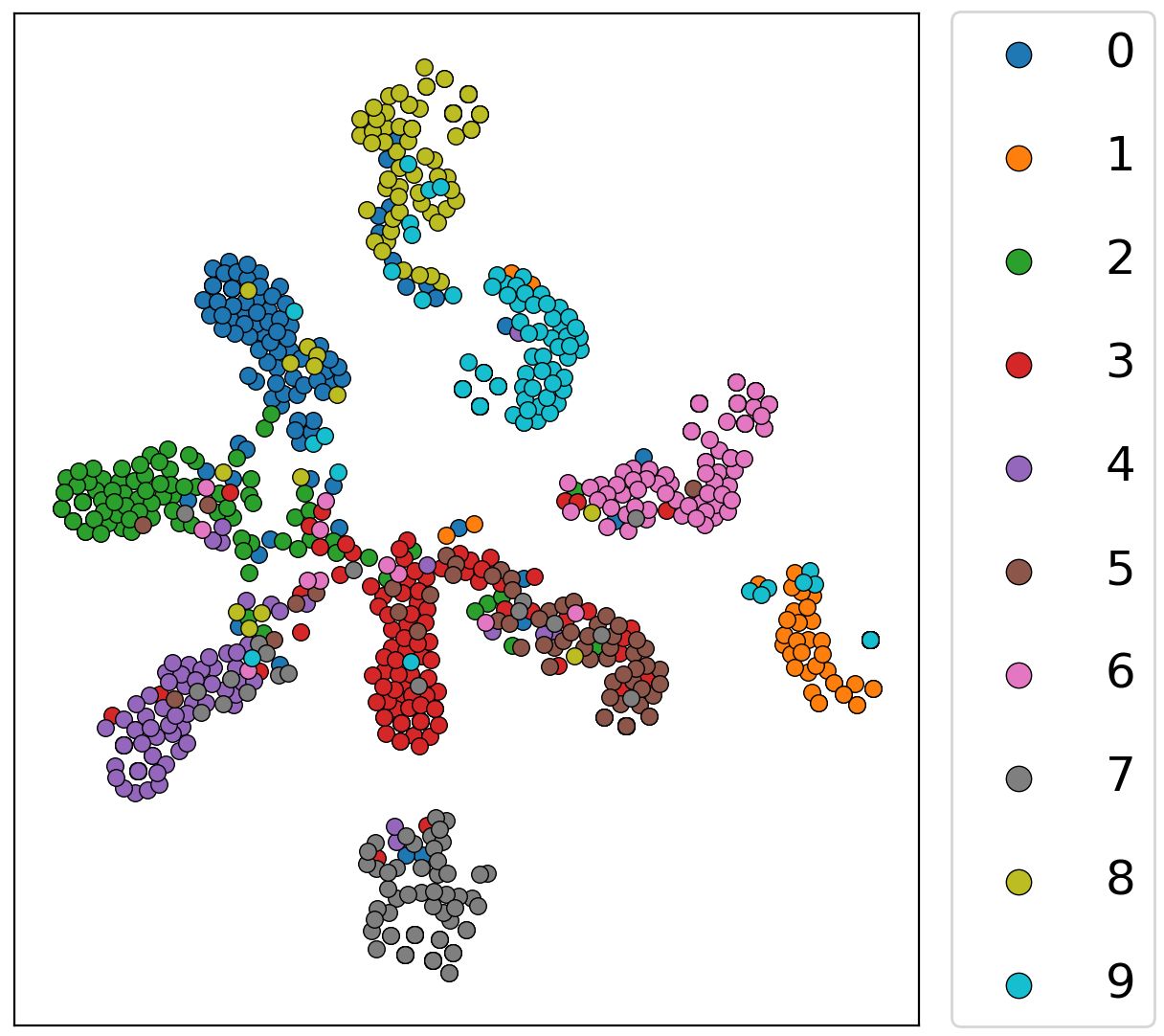} \\{(a) First ResBlock}
 \end{minipage} \
 \begin{minipage}{0.47\linewidth}
    \centering
    \includegraphics[width=1\linewidth]{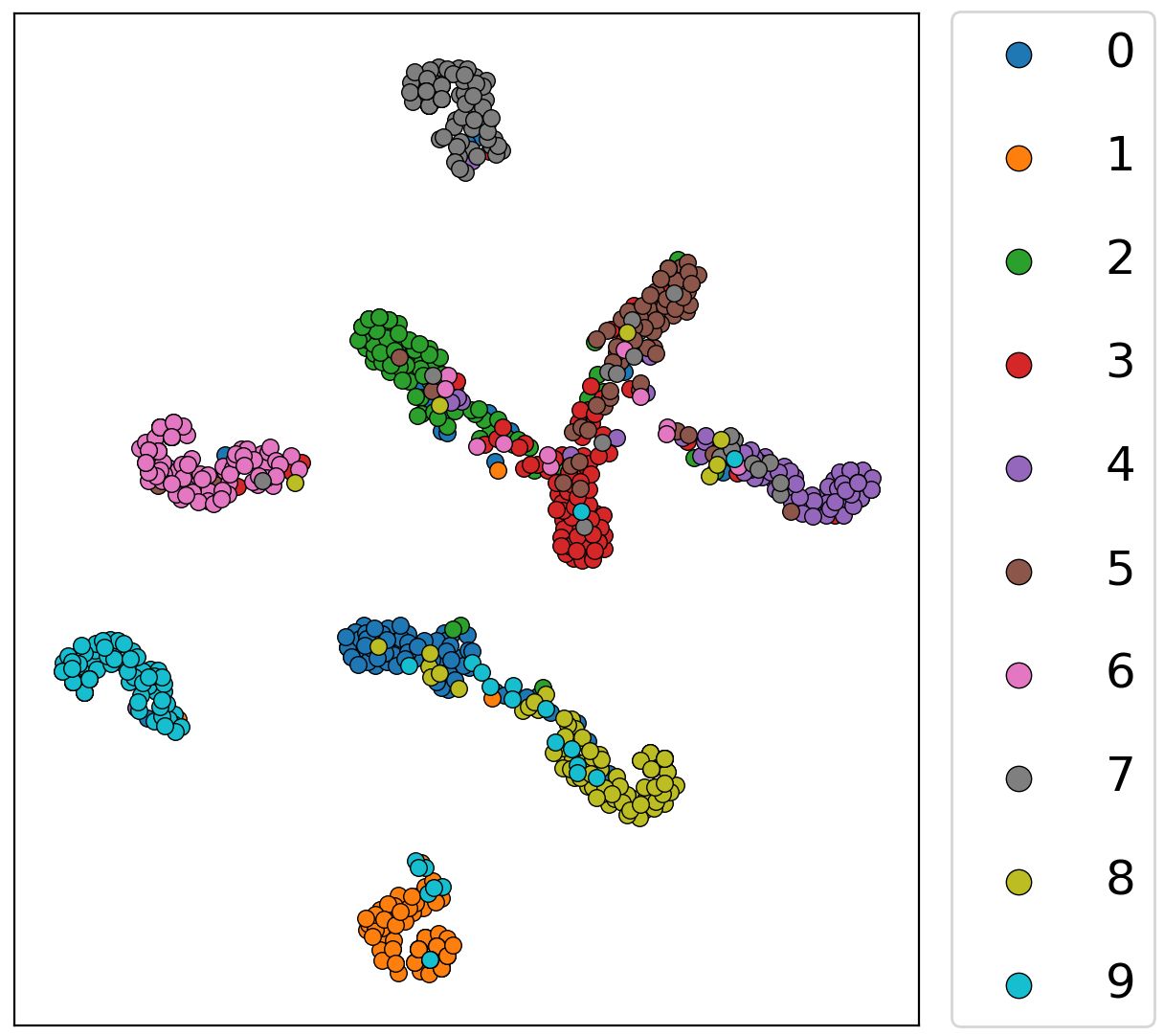} \\{(b) Last ResBlock}
 \end{minipage}\vspace{-1mm}
   \caption{t-SNE \cite{maaten2008visualizing} visualization of the distribution of the modulation parameter vector $\bm\gamma$ lying at the first and last ResBlock of ResNet-32 trained on long-tailed CIFAR-10.}
   \vspace{-1mm}
  \label{fig_tsne}
 \end{minipage}
\end{figure}

\textbf{Visualization of modulation parameters:} To give an explanation on how the proposed modulator network contributes to long-tailed tasks, we find patterns in the learned $\bm\gamma$ and $\bm\beta$ with t-SNE \cite{maaten2008visualizing}. Concretely, we modulate the last convolutional layer of each ResBlock for ResNet-32 and train the network on long-tailed CIFAR-10. Once the network is trained, we visualize the distribution of $\bm \gamma$ vectors from one layer in 2D for a total of 1000 validation samples (please refer to supplementary materials for visualization of $\bm \beta$). From Fig. \ref{fig_tsne}, it is clear that: 1) The modulation parameters contain traits of the training dataset, since the learned parameters are grouped according to the classes, which indicates the modulator network class-wisely affect the classification network in the feature space; 2) Both shallow and deep layers of the classifier can be influenced by the modulator network, which shows that the modulator network can transfer the meta information without an architectural prior.

In reality, our MFM can be viewed as a generalized BN, where the parameters are learned by meta knowledge from an unbiased meta set, and the learned modulator network is a meta-learner capable of modeling the difference between training and meta distribution. In a word, the modulator network acts as a modulator re-balancing the contribution of each class during training, which inspires us to apply  the learned modulator network to train another class-imbalanced dataset (See in Section 4.3).

\begin{table} [t]
\caption{Ablation analysis for various modulation manners. The models are trained on long-tailed CIFAR-10 with an imbalance factor of 100 and top-1 error rates are reported.
}\vspace{-2mm}
\centering
\setlength{\tabcolsep}{2.5mm}
\begin{tabular}{c c c c|c}
  \toprule
  Model&ResBlock1&ResBlock2&ResBlock3&Error rate\\
  \hline
  BaseModel& & & &26.90\\
  MFM&\checkmark & & &26.23\\
  MFM& &\checkmark& &23.98\\
  MFM& & &\checkmark&19.72\\
  MFM&\checkmark&\checkmark&\checkmark & 21.71\\
  \hline
  MFM (with $\bm\beta :=\bm 0$)&&&\checkmark& 21.28\\
  MFM (with $\bm\gamma:=\bm 1$)&&&\checkmark& 21.61\\
  \bottomrule
  \end{tabular}
\label{table_ablation}
\vspace{-5mm}
\end{table}
\textbf{Ablation analysis:} For better understanding of the proposed MFM model, we conduct an ablation study on long-tailed CIFAR-10 with an imbalance factor of 100. We firstly quantify the effectiveness of modulator network implemented on different layers of classification network. As shown in Table \ref{table_ablation}, modulating deep or shallow layers of classification network has a significant performance gain compared with BaseModel, which complies with the findings shown in Fig. \ref{fig_tsne} that the meta information can be transferred to classification network even by one layer. Beyond that, we find that modulating the last ResBlock is superior to other ResBlocks, including all three ResBlocks used together. This is probably because the shallow convolution layers capture simple and general visual textures, whereas deep layers pay more attention to semantic and class-specific patterns \cite{zeiler2014visualizing}. As shown in Fig. \ref{fig_tsne}, the modulator network of MFM seems to learn class-based modularity for transferring the meta information, thus modulating deep layers is superior to shallow layers. To corroborate the aforementioned findings are architecture-agnostic, we as well conduct an ablation study with a modified LeNet \cite{lecun1998gradient} on long-tailed Fashion-MNIST \cite{xiao2017} in supplementary materials.

We further study the effect of scale factor $\bm\gamma$ and bias factor $\bm\beta$ on classification accuracy. Concretely, we separately train two models: 1) with learnable $\bm\gamma$ and constant $\bm\beta=\bm 0$; 2) with learnable $\bm\beta$ and constant $\bm\gamma=\bm 1$. The results in Table \ref{table_ablation} show that both $\bm\gamma$ and $\bm\beta$ plays important roles in our MFM model, since there would be a performance drop once either of them is fixed.

\subsection{Modulator network transfer across datasets}
\begin{table}[t]
\caption{Top-1 error rates for transfer modulator network between long-tailed Fashion-MNIST and CIFAR-10. For Fashion-MNIST, we denote \textbf F as BaseModel, \textbf F$\rightarrow$ \textbf F as MFM, \textbf C $\rightarrow$ \textbf F as MFM with a fixed modulator network transferred from CIFAR-10, respectively, so as to CIFAR-10.}\vspace{-2mm}
\label{table_transfer}
\begin{center}
\begin{tabular}{c|c|c|c|c}
\toprule
Imbalance factor& 200& 100& 50& 10 \\
\midrule
\textbf F&16.12$\pm$0.71 &14.01$\pm$0.26 &12.40$\pm$0.13 &9.29$\pm$0.15\\
\textbf F$\rightarrow$\textbf F&14.37$\pm$0.41 &12.12$\pm$0.30 &11.44$\pm$0.60 &9.59$\pm$0.40 \\
\textbf C$\rightarrow$\textbf F&13.89$\pm$0.39 &12.79$\pm$0.35 &11.54$\pm$0.26 &9.23$\pm$0.19 \\
\hline
\hline
\textbf C &32.38$\pm$0.65 &26.90$\pm$0.72 &22.31$\pm$0.73 &13.55$\pm$0.46 \\
\textbf C$\rightarrow$\textbf C &24.93$\pm$0.28 &19.72$\pm$0.46 &16.16$\pm$0.32 &11.07$\pm$0.38 \\
\textbf F$\rightarrow$\textbf C &24.35$\pm$1.37 &20.40$\pm$0.99 &16.91$\pm$0.78 &12.67$\pm$0.68 \\
\bottomrule
\end{tabular}
\end{center}
\vspace{-3mm}
\end{table}

As aforementioned, to minimize the empirical risk of meta-dataset during training, the modulator
aims to re-balance the contribution of each class. As we explicitly learn a modulator network satisfying to rebalance the classes conditioned on the input soft label, a natural conjecture is that wether we can transfer the learned meta modulator network to train another class-imbalanced dataset. To verify this, we experiment with long-tailed versions of CIFAR-10 and Fashion-MNIST \cite{xiao2017} (abbreviated as \textbf{C} and \textbf{F} for brevity) which involves two distinct domains and are created by the same strategy described in Section 4.1. We train ResNet-32 \cite{he2016deep} with the same settings in Section 4.1 on one dataset under an imbalance factor of 200, and adapt the learned modulator network to train the other one under four different imbalance factors: 10, 50, 100 and 200. The results are listed in Table \ref{table_transfer}. It can be demonstrated that: 1) For \textbf{F} dataset, transferring the learned modulator net from \textbf{C} (\textbf{C} $\rightarrow$ \textbf{F}), consistently achieving obvious improvements compared with BaseModel directly trained on \textbf{F} and competitive results compared with MFM with an updating modulator network on \textbf{F} during training (\textbf{F} $\rightarrow$ \textbf{F}), and the similar findings to \textbf{C} dataset; 2) Fixing the imbalance factor to train the modulator network, the transfer results on the other dataset under different imbalance factors consistently outperform the BaseModel, which indicates the learned modulator network is robust to the change in the number across different classes of target dataset.
\subsection{Experimental for long-tailed scene recognition}

\begin{table}[t]
\caption{Top-1 error rates of all comparison methods on MIT-67 scene dataset.}\vspace{-3mm}
\label{table_mit}
\begin{center}
\begin{tabular}{c|c|c|c|c|c}
\toprule
\#& Method& Error rate& \#& Method& Error rate \\
\midrule
1& BaseModel& 28.54 &5&L2RW \cite{ren2018learning}& 26.50\\
2& Focal \cite{lin2017focal}& 28.21 &6&MW-Net \cite{shu2019meta}&26.24\\
3& CB \cite{cui2019class} &24.98 &7&Ours& 24.63\\
4& CoSen CNN \cite{Khan2017Cost}& 26.80 &8&Ours (with WH-layer)&\textbf{24.20}\\
\bottomrule
\end{tabular}
\end{center}
\vspace{-5mm}
\end{table}

Indoor scene recognition is a challenging problem where visual phenomena naturally follows a skewed distribution. We experiment with MIT-67 scene dataset \cite{quattoni2009recognizing}, containing 15,620 images belonging to 67 classes with varied class number between 101 and 738. We use the complete dataset with imbalanced train/test splits of 60\%/40\% following the setting in \cite{Khan2017Cost}, and a pre-trained VGG-11 \cite{simonyan2014very} (only reserving the last FC layer and adding an average-pooling before it) trained with cross-entropy loss as our BaseModel. We randomly copy 20 samples per class from training set as meta data, and directly modulate the features of the last convolution layer. Please refer to supplementary materials for more details about the training. Table \ref{table_mit} summarizes top-1 test error rates, and our MFM outperforms the BaseModel by 4.34\% and achieves the best performance among all comparison methods.

\section{Conclusion}
In this work we have proposed a meta-learning framework to address the long-tailed recognition problem through directly modulating the features of classification network. Our key insight is to train the classification network on long-tailed training data, such that adapting it to finely fit the implicit class-balanced label distribution. We further introduce a modulator network to generate the modulation parameters, such a pre-learned modulator can be transferred to train the classification network across different long-tailed datasets. Experimental results on benchmark datasets demonstrate the superiority of the proposed method in long-tailed recognition tasks. As our meta feature modulator is architecture-agnostic, such an adaptive feature manipulation method can be readily applied to
other machine learning tasks, like adversarial attacks and transfer learning.

\bibliographystyle{unsrt}
\bibliography{mybib}

\newpage
\appendix
\renewcommand\thesection{\Alph{section}}
\section{Modulator network}
\subsection{Weight hashing}
Weight hashing \cite{chen2015compressing} is a mapping that maps a low-dimensional vector $\bm x \in \Re^ m$ to a high-dimensional one $\bm\theta \in \Re^d$ by hash tricks, where $d > m$ is commonly satisfied. Specifically, weight hashing replicates each element of $\bm x$ in multiple locations of $\bm\theta$ and randomly flips its sign to reduce the covariance of copied coefficients. Formally, for the $i$-th coefficient of $\bm\theta$, we have
\begin{equation*}
\bm\theta(i)=\bm x(\kappa(i))\zeta(i),
\end{equation*}
where $\kappa(i)\in\{1,2,...,m\}$ and $\zeta(i)\in\{1,-1\}$ are hashing functions determined randomly. Weight hashing can be efficiently implemented as a fully-connected layer with fixed weights \cite{shaban2017one} as follows:
\begin{align*}
\bm\theta &= W\bm x,  \\
\mathrm{s.t.}\quad W(i,j)&= \zeta(i)\delta_j((\kappa(i))),
\end{align*}
where $\delta_j(\cdot)$ is discreet Dirac delta function. Note that this fully-connected layer is parameter-free, since the weight matrix $W$ is preset by random hashing functions before training and kept fixed during both the training and inference phases.

\subsection{Modulator network with weight hashing layer}
\begin{figure}[t] \vspace{0mm}
  \centering
    \includegraphics[width=0.45\textwidth]{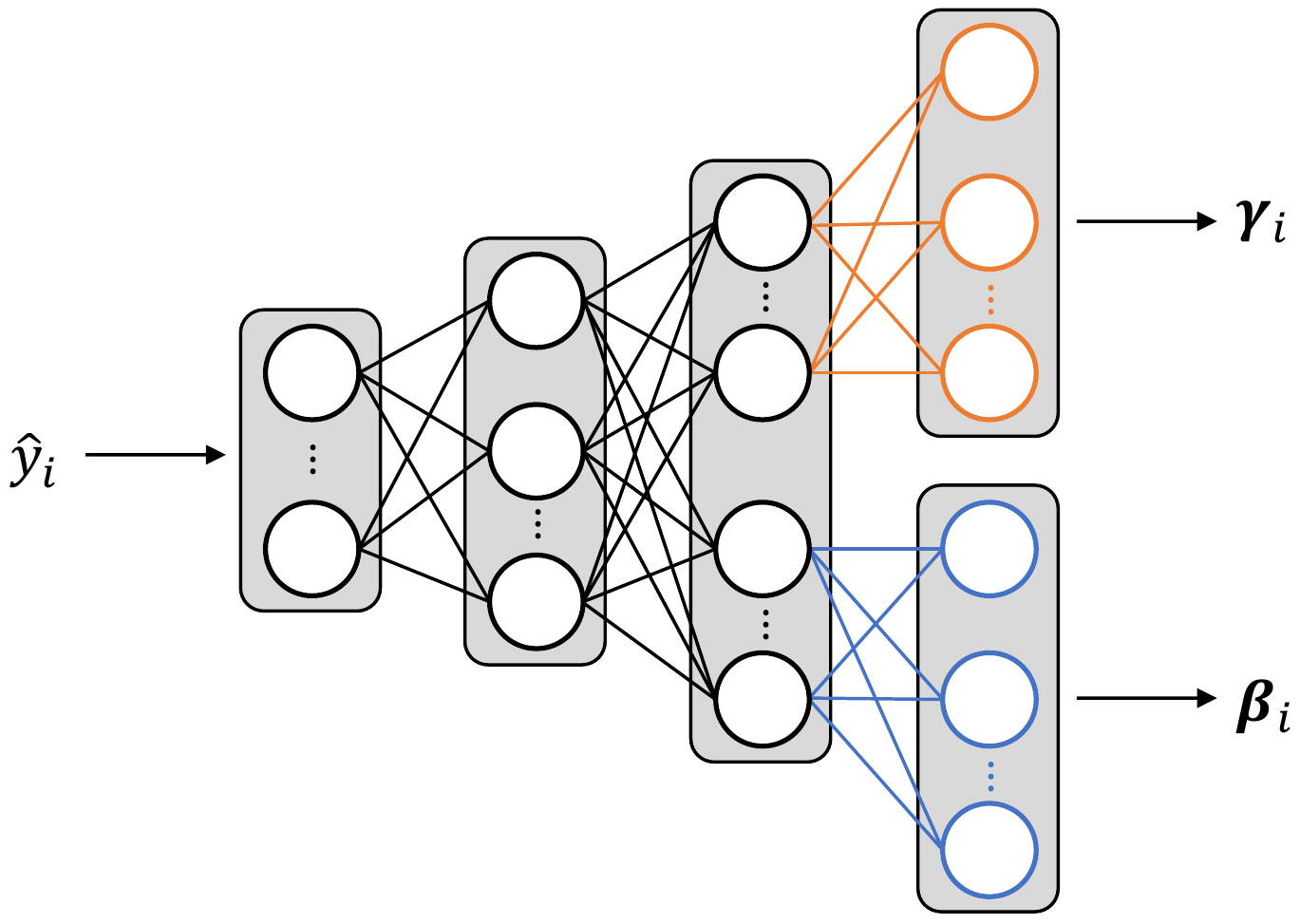}
  \caption{Architecture of the proposed modulator network with weight hashing layer. This modulator network can be roughly divided into two parts: 1) The first three layers, which make up a standard MLP with one hidden layer to output compressed vector $\hat{\bm\gamma}$ and $\hat{\bm\beta}$ for $\bm\gamma$ and $\bm\beta$, and 2) The last weight hashing layer, which separately decompresses $\hat{\bm\gamma}$ and $\hat{\bm\beta}$ to $\bm\gamma$ and $\bm\beta$ by two weight hashing mappings.}
  \label{fig_modulator}
\end{figure}
As mentioned in Section 3.2 of the main text, for the $i$-th training sample, our modulator network adopts a single multilayer perceptron (MLP) architecture with one hidden layer to map its soft label $\hat y_i$ to the expected modulation parameters $\{\bm\gamma_i,\bm\beta_i\}$. However, the dimension of the vector concatenated by $\bm\gamma_i$ and $\bm\beta_i$ are probably very high, even if we just modulate features of a wide network layer, such as the 4096-neuron fully-connected layer of VGG-19 \cite{simonyan2014very} and 2048-channel convolution layer of ResNet-50 \cite{he2009learning}, which might arise over-fitting. Against this issue, we introduce an additional weight hashing layer (WH-layer), and the resulting modulator network is shown in Fig. \ref{fig_modulator}. The WH-layer consists of two hash mappings, which map a low-dimensional vector $\hat{\bm\gamma_i}$ (or $\hat{\bm\beta_i}$) to desired $\bm\gamma_i$ (or $\bm\beta_i$). An intuitive explanation is that the MLP produces compressed vector $\hat{\bm\gamma_i}$ and $\hat{\bm\beta_i}$ (dimidiated from one single vector) for $\bm\gamma_i$ and $\bm\beta_i$, and the weight hashing layer separately decompresses them to the desired $\bm\gamma_i$ and $\bm\beta_i$.

In Section 4.3 of the main text, we have mentioned that the modulator network can conveniently extend the modulation operation to more strategies, like FiLM \cite{perez2018film} and attention-based modulation \cite{mnih2014recurrent, vaswani2017attention}. Here we give several types of modulation methods. 1) FiLM based method, which channel-wisely modulates the features by affine transformation, i.e., $\bm\gamma$ and $\bm\beta$ are obtained by the modulator network, and not constrained with any activation function. 2) Channel attention \cite{chen2017sca,hu2018squeeze} based method, which is easily implemented by fixing $\bm\beta_i$ as a constant $\bm 0$ vector, and constraining $\bm\gamma$ by activation function $\mathrm{sigmoid}(\cdot)$. 3) Gated attention \cite{mnih2014recurrent, vaswani2017attention} based method, which constrains $\bm\beta$ to $\bm 0$ vector and $\bm\gamma$ with activation function $\mathrm{softmax}(\cdot)$. In fact, we can think that constraining the modulation parameters with different activation functions correspond to different modulation methods. In this paper, we empirically adopt $\bm\gamma$ constrained with $C\times\mathrm{softmax}(\cdot)$ (for layer-wise feature) and $\bm\beta$ without constrains to modulate the intermediate features of classification networks, where $C$ is channel number of the associated layer.

\section{Implementation details}
\begin{figure}[t]
  \begin{minipage}{0.33\linewidth}
  \centering
  \includegraphics[width=1.0\textwidth]{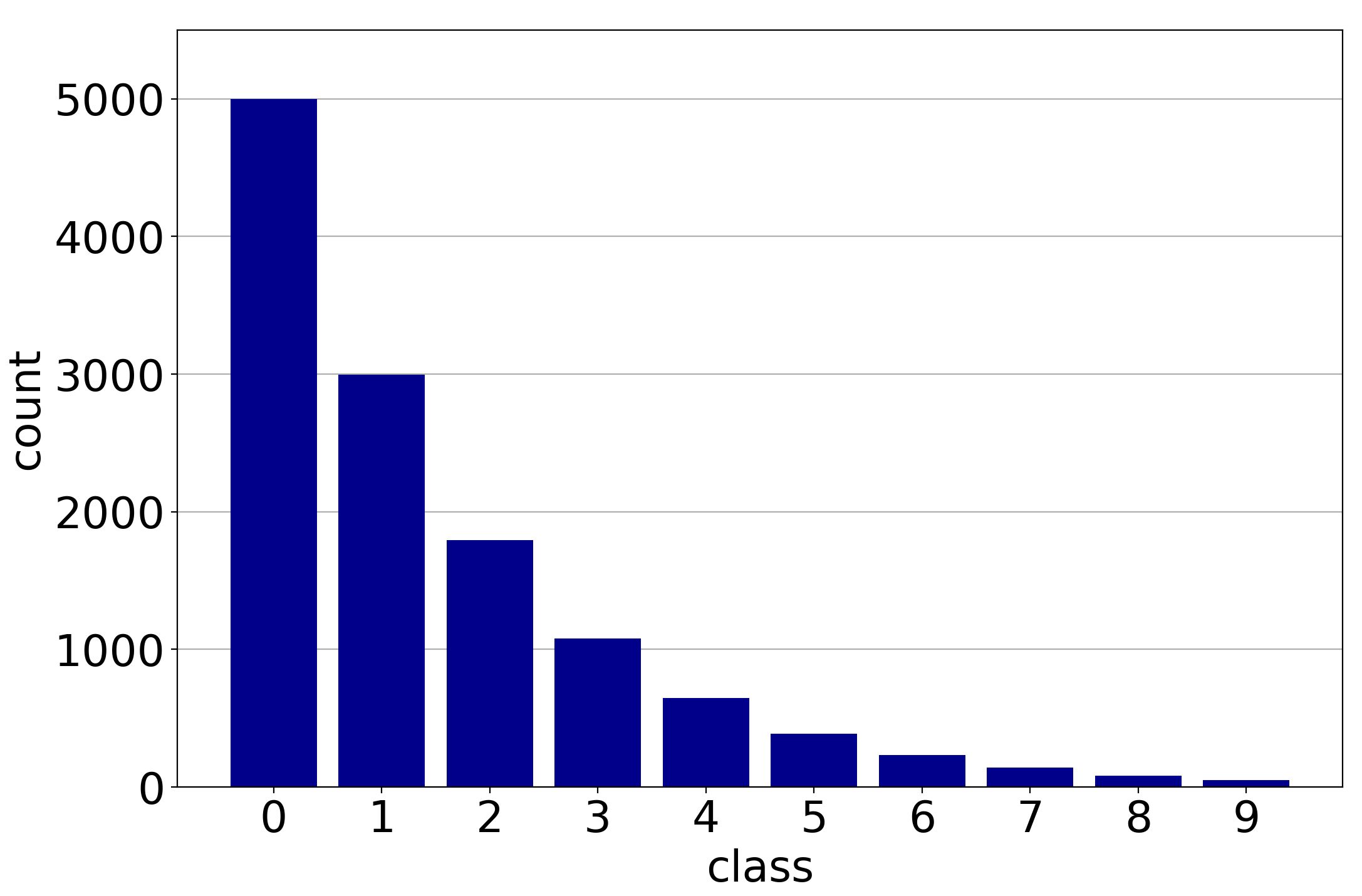}\\{(a)}
  \end{minipage}
  \begin{minipage}{0.33\linewidth}
  \centering
  \includegraphics[width=1.0\textwidth]{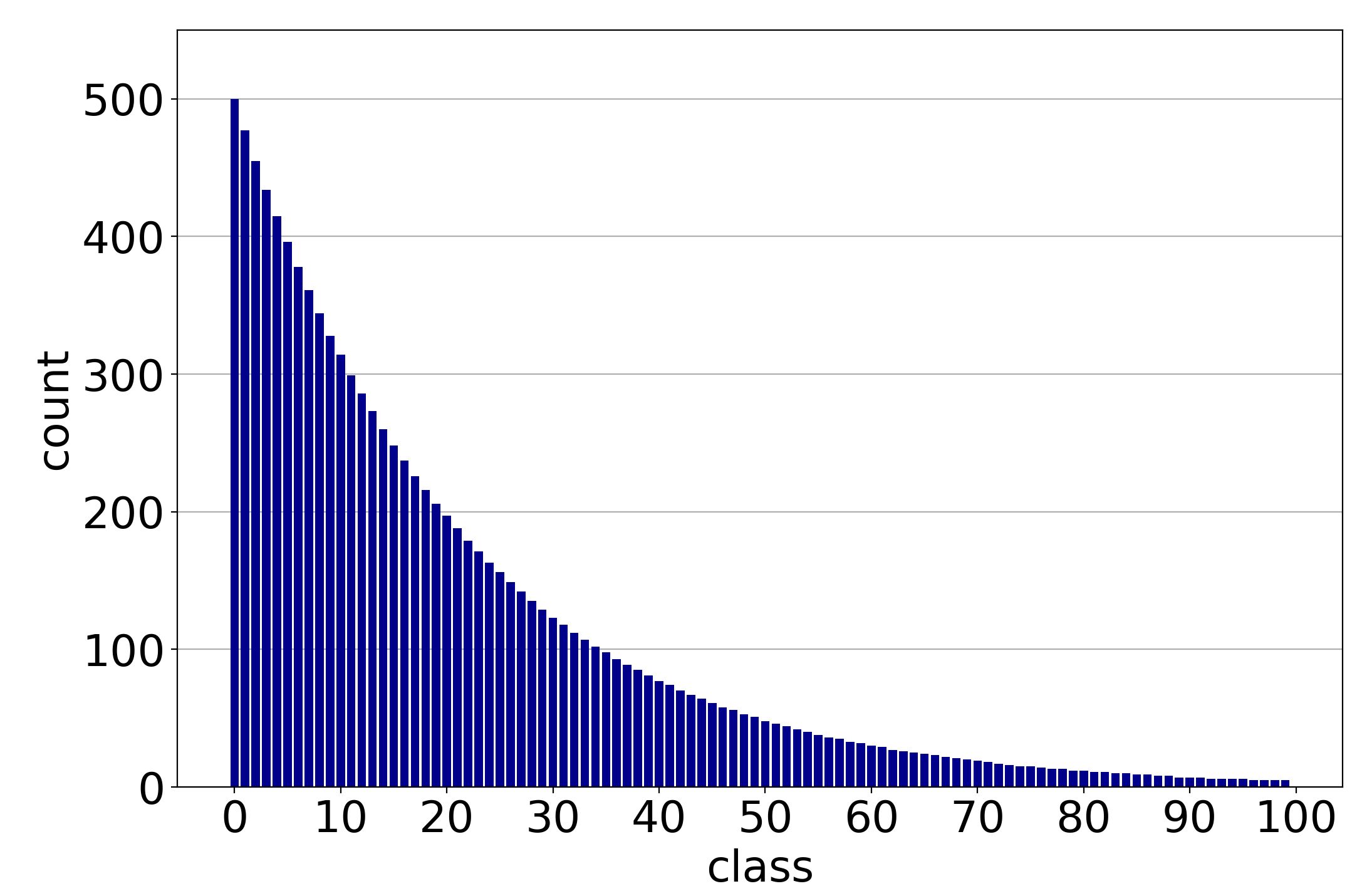}\\{(b)}
  \end{minipage}
  \begin{minipage}{0.33\linewidth}
  \centering
  \includegraphics[width=1.0\textwidth]{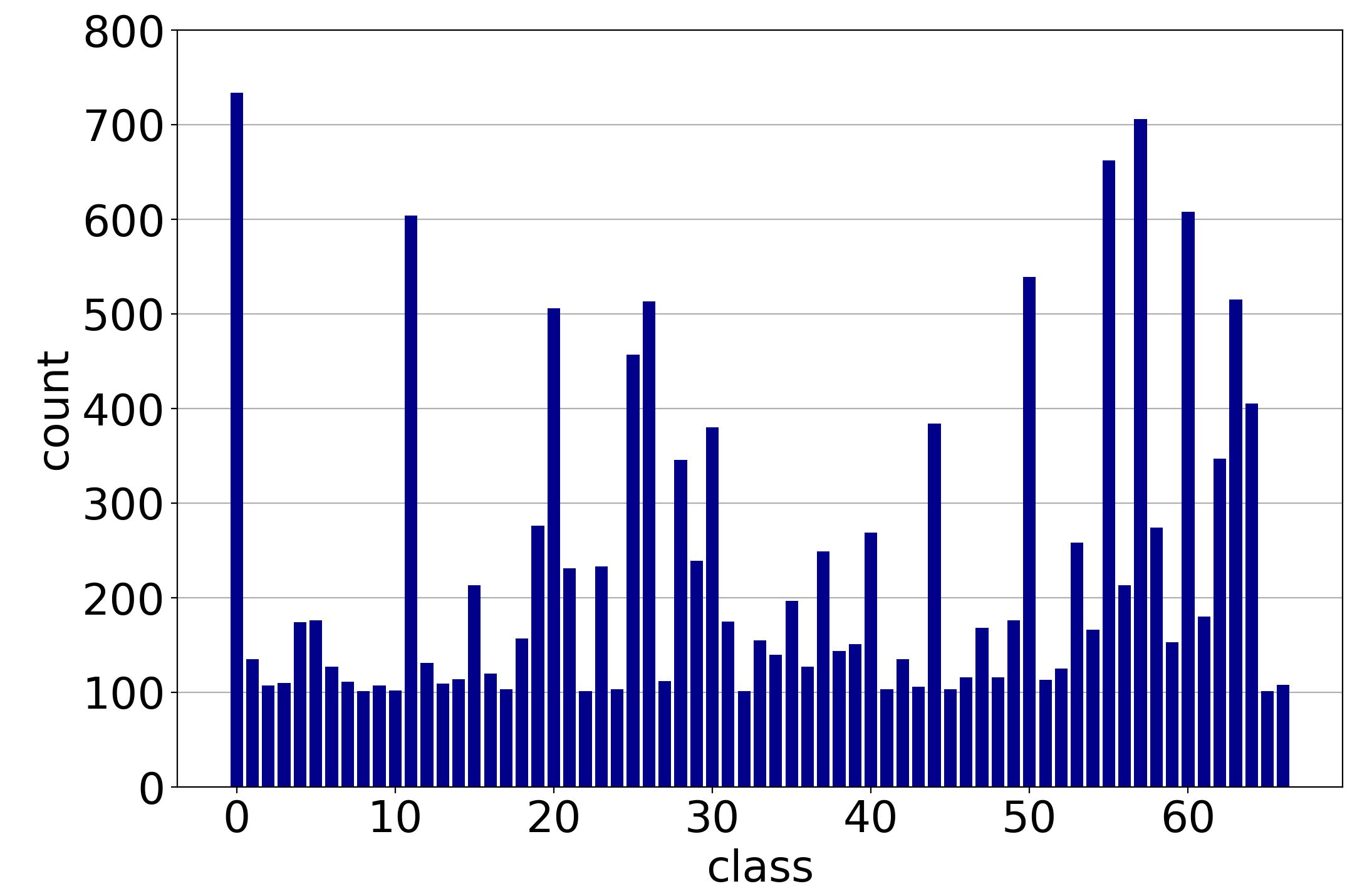}\\{(c)}
  \end{minipage}
  \caption{Number of training examples per class in various typical datasets: (a) long-tailed CIFAR-10 with an imbalance factor of 100; (b) long-tailed CIFAR-100 with an imbalance factor of 100; (c) MIT-67 scene dataset.}
  \label{fig_train_distribution}
\end{figure}
\textbf{Label distributions.} In Fig. \ref{fig_train_distribution}, we visualize some example distributions of long-tailed CIFAR and MIT-67 scene datasets used in Section 4.

\textbf{Implementation details for CIFAR.} For CIFAR-10 and CIFAR-100, we follow the data
augmentation strategies proposed in \cite{he2016deep} for training: 4 pixels are padded on each side, and a 32$\times$32 crop is randomly sampled from the padded image or its horizontal flip. The proposed MFM is trained with SGD with an initial learning rate of 0.1, a momentum of 0.9, a weight decay of $5\times10^{-4}$ with mini-batch size of 100 on one GPU of NVIDIA 1080Ti for a total of 200 epochs, and the learning rate of modulator network (an MLP with an 100-neuron hidden layer) is fixed as $1\times10^{-3}$. We decay the learning rate by 0.1 at the 160-$th$ epoch and again at the 180-$th$ epoch.

\textbf{Implementation details for MIT-67 scene dataset.} For MIT-67 scene dataset, we adopt a simple data augmentation: Each image is resized as 224$\times$224, and then randomly flipped horizontally. The proposed MFM is trained by SGD with a initial learning rate of 0.01, a momentum of 0.9, a weight decay of $1\times10^{-4}$ with mini-batch size of 40 on two GPUs of NVIDIA 1080Ti for a total of 90 epochs, and the learning rate is decayed by 0.1 every 30 epochs. For the modulator network, we set the neuron number of hidden layer as 256, and fix the learning rate as $1\times10^{-4}$. We as well evaluate the proposed modulator network with weight hashing layer. For $\bm\gamma$, we set the dimension of its compressed vectors $\hat{\bm\gamma}$ as 256, and so does for $\bm\beta$.

\section{Additional results}
\subsection{Imbalanced test label distributions}

\begin{figure}[t]
  \begin{minipage}{0.33\linewidth}
  \centering
  \includegraphics[width=1.0\textwidth]{train_cifar10_100.jpg}\\{(a) Train}
  \end{minipage}
  \begin{minipage}{0.33\linewidth}
  \centering
  \includegraphics[width=1.0\textwidth]{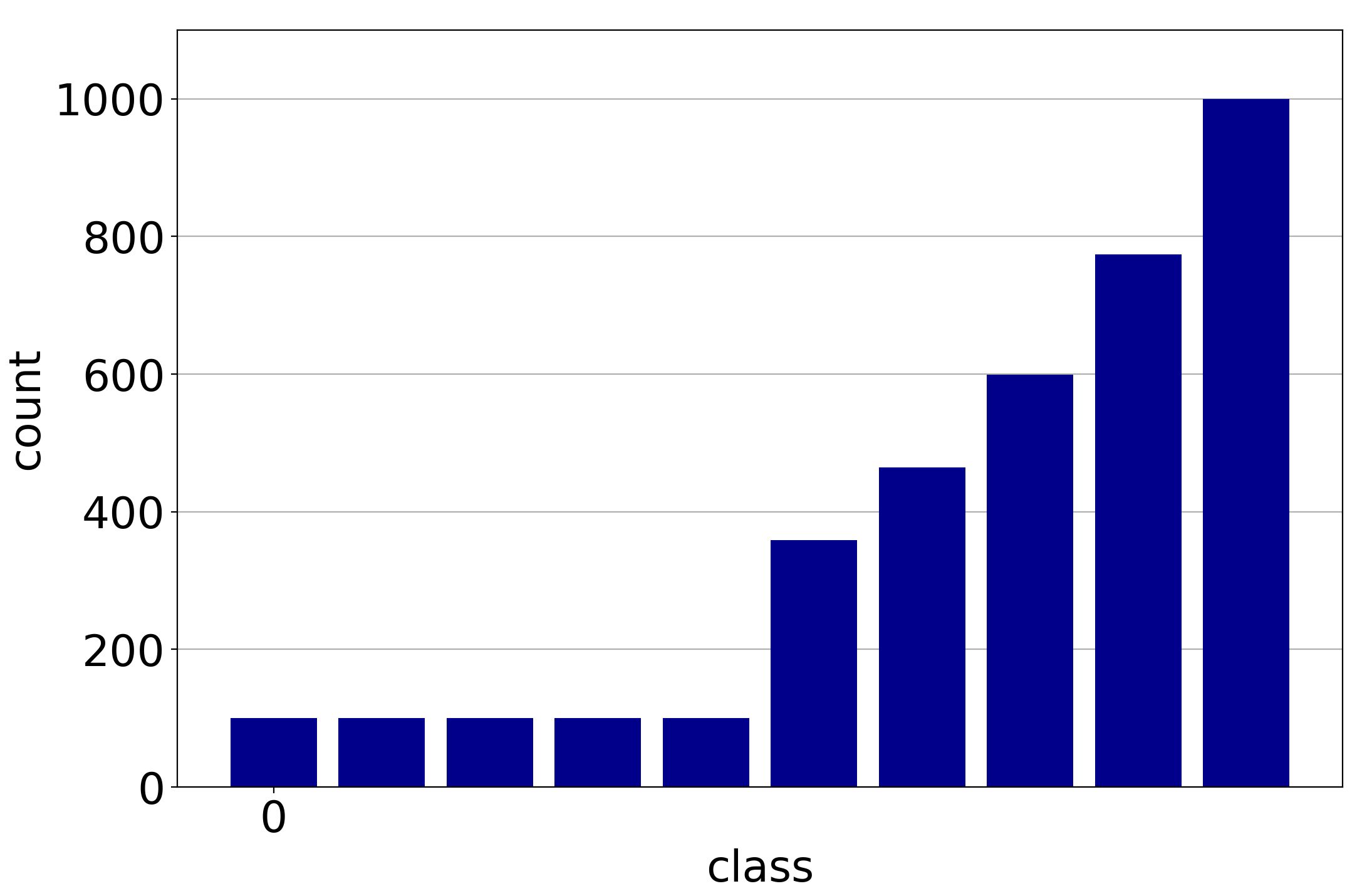}\\{(b) Test-1}
  \end{minipage}
  \begin{minipage}{0.33\linewidth}
  \centering
  \includegraphics[width=1.0\textwidth]{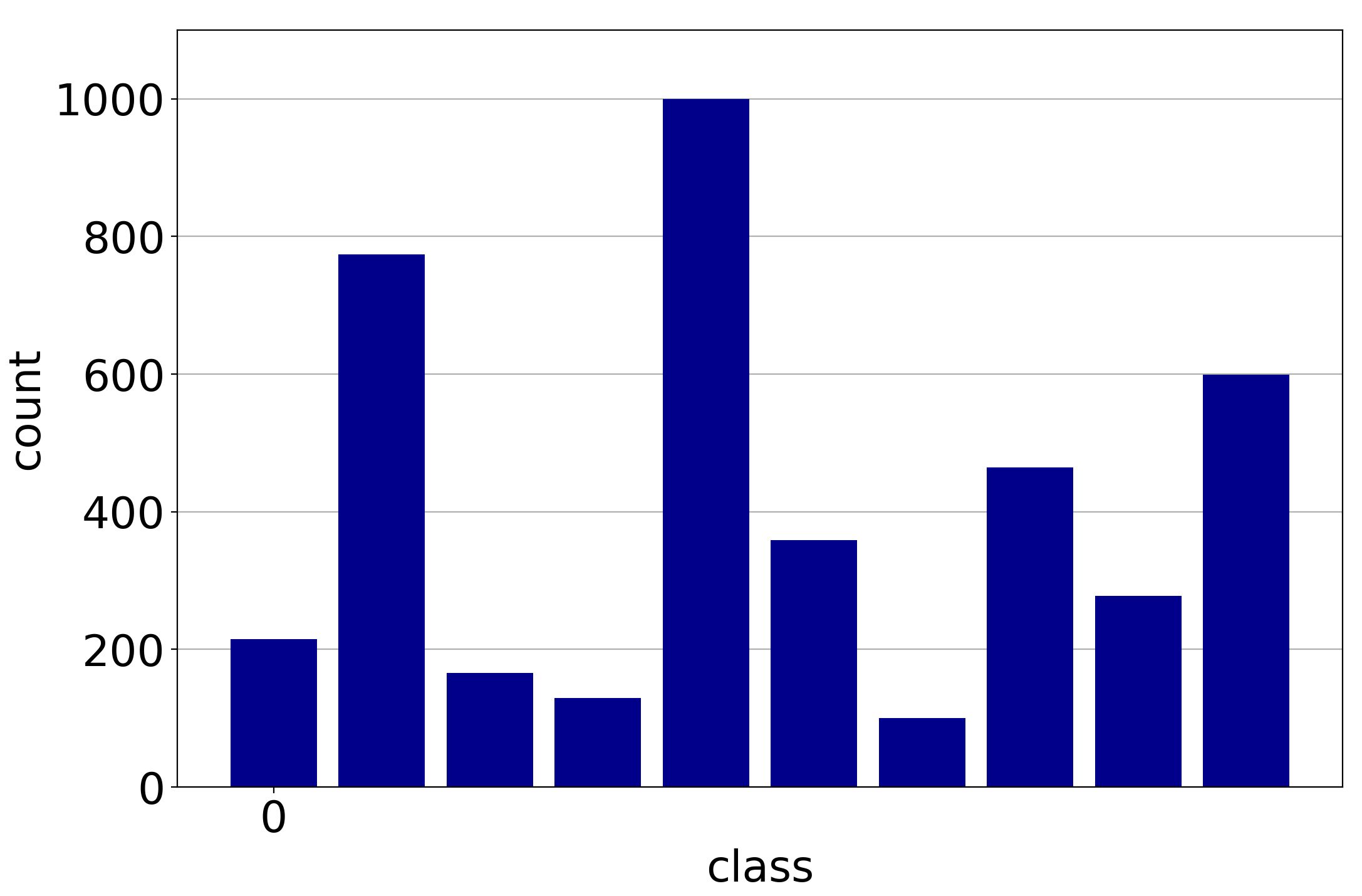}\\{(c) Test-2}
  \end{minipage}
  \caption{Example distributions when train and test distributions are both imbalanced. The training set is long-tailed CIFAR-10 with an imbalance factor of 100, Test-1 and Test-2 sets are created by the strategy described in Section C.1}
  \label{fig_test_distribution}
\end{figure}
The majority of our experiments have conducted on the test set with  uniform test distribution setting, and  the results in Section 4 of the main text substantiate the superiority of our method in these scenarios. Here we test with the sets with imbalanced label distributions. To make the test label distribution significantly different from the training one, we sample from the original test set and create two specific test sets. We firstly use the same rule as described in Section 4 to generate imbalanced test label distribution with an imbalance of 10, and then for 1) Test-1: reversing the frequency of the labels and truncate the first five class have the same number of samples. 2) Test-2: permuting randomly the frequency of the labels. Fig. \ref{fig_test_distribution}(b)-(c) present the label distributions of the two test sets.

We train the model under three different imbalance factors: 10, 50, 100 and evaluate the performance on Test-1 and Test-2 sets, respectively. The results are listed in Table \ref{table_imb_test}. It is clear that: 1) When the training and test sets are both imbalanced, our MFM as well consistently outperforms BaseModel and MW-Net under all experimental settings. 2) Our MFM tends to achieve more performance gain when the training set becomes more imbalanced, e.g., training on the long-tailed CIFAR-10 with an imbalanced factor of 100, it achieves 20.81\% error rate on Test-1, which is over 18.96\% lower than that of MW-Net in such an extremely imbalanced scenario.

\begin{table}[t]
\caption{Top-1 error rates of BaseModel, MW-Net and our MFM evaluated on Test-1 and Test-2. The best results are highlighted in \textbf{bold}.}
  \label{table_imb_test}
  \centering
  \begin{tabular}{c|c|c|c|c|c|c}
  \toprule
  Test set & \multicolumn{3}{c|}{Test1} & \multicolumn{3}{c}{Test2}\\
  \hline
  Training imbalance & 100 & 50 & 10 & 100 & 50 & 10 \\
  \hline
  BaseModel & 42.13&28.03 &15.34 &29.53 &19.85 &12.19\\
  MW-Net & 39.77& 31.11&15.72 & 25.95& 20.37 & 12.88\\
  Ours  & \textbf{20.81}&\textbf{15.58} & \textbf{9.85}& \textbf{18.90}& \textbf{16.01}& \textbf{9.62}\\
  \bottomrule
  \end{tabular}
\end{table}
\subsection{Visualization of shifting modulation parameters}
\begin{figure}[h]
  \begin{minipage}{0.48\linewidth}
  \centering
  \includegraphics[width=0.65\textwidth]{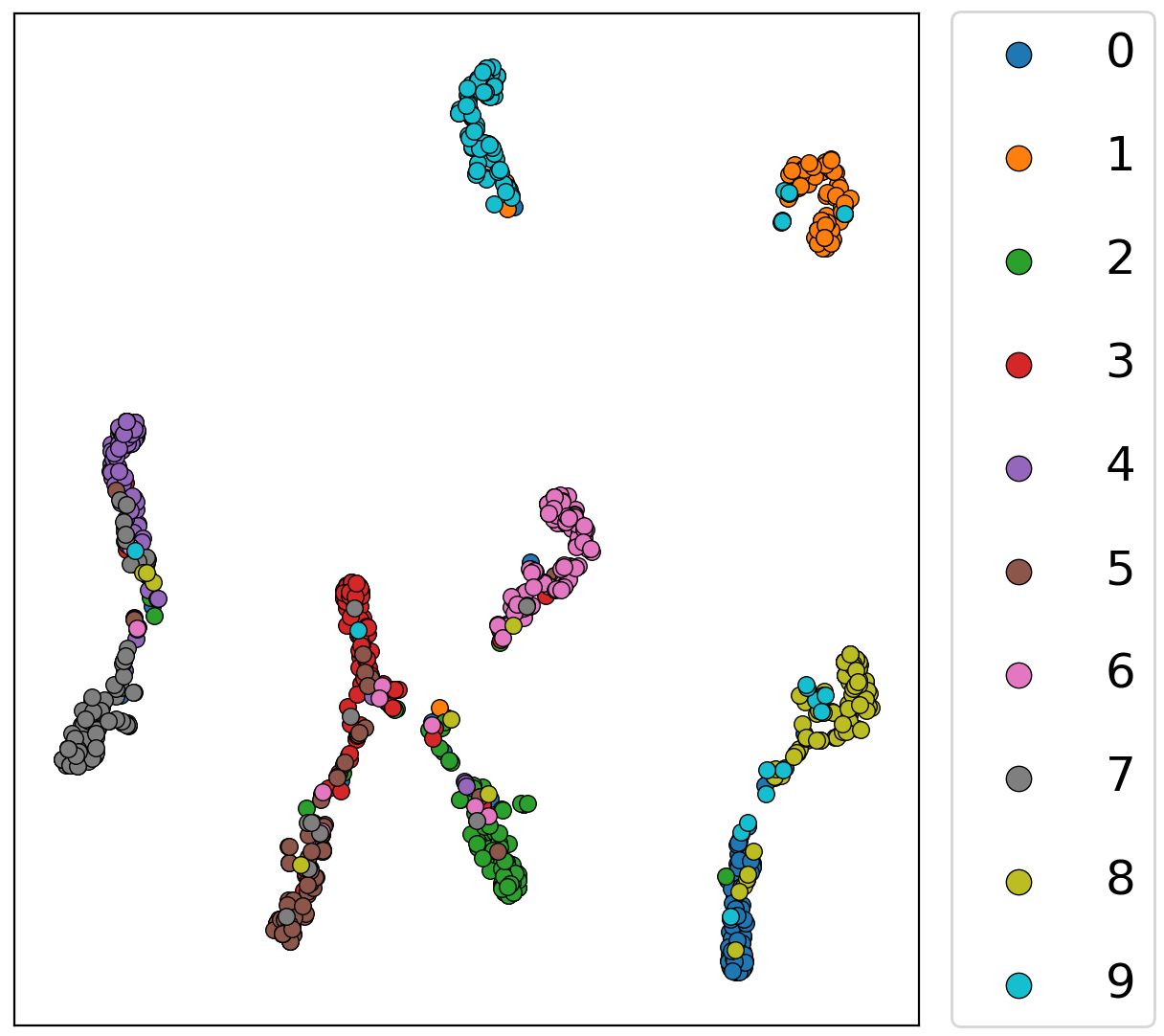}\\{(a) First ResBlock}
  \end{minipage}
  \begin{minipage}{0.48\linewidth}
  \centering
  \includegraphics[width=0.65\textwidth]{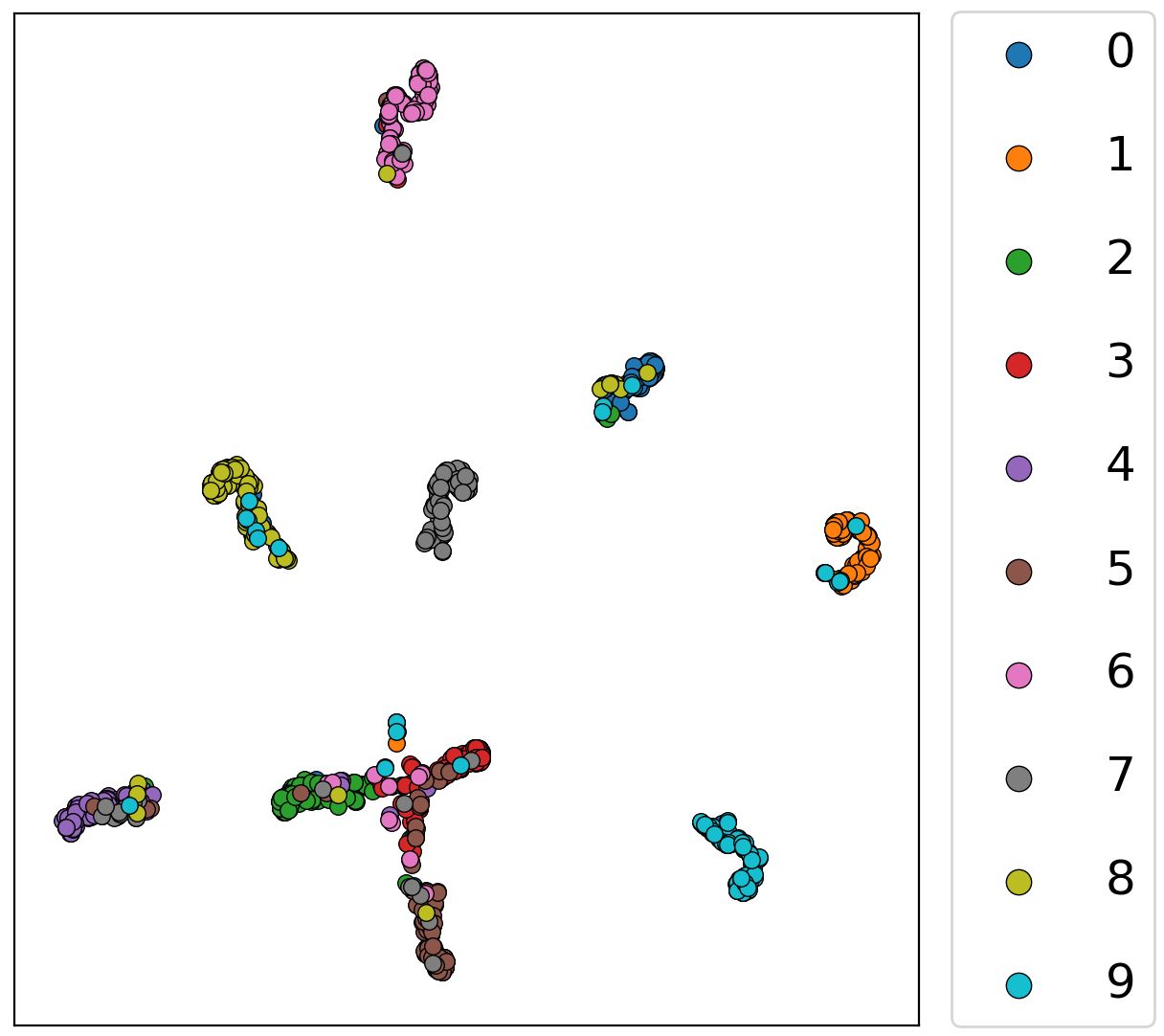}\\{(b) Last ResBlock}
  \end{minipage}
  \caption{t-SNE \cite{maaten2008visualizing} visualization of the distribution of the modulation parameter vector $\bm\beta$ lying at the first and last ResBlock of ResNet-32 trained on long-tailed CIFAR-10.}
  \label{fig_visualize_beta}
\end{figure}

In Section 4.2 of the main text, we have layer-wisely visualized the distribution of $\bm\gamma$ vector of each validation samples through t-SNE \cite{maaten2008visualizing} to understand the modulator network. Here we further visualize the distribution of $\bm\beta$ vector in Fig. \ref{fig_visualize_beta}. As shown, the learned shifting modulation parameters $\bm\beta$ are grouped according to the classes at both the shallow and deep layers, which further confirms the findings we have mentioned in Section 4.2 of the main text.

\subsection{Ablation analysis on modified LeNet}
\begin{table} [h]
\caption{Ablation analysis for various modulation manners. The models are trained on long-tailed Fashion-MNIST with an imbalance factor of 1000 and top-1 error rates are reported.
}\vspace{0mm}
\centering
\setlength{\tabcolsep}{2.5mm}
\begin{tabular}{c c c c c|c}
  \toprule
  Model&Conv1&Conv2&Conv3&FC1&Error rate\\
  \hline
  BaseModel& & & & &26.24\\
  MFM&\checkmark & & & & 24.67\\
  MFM& &\checkmark& & & 23.36\\
  MFM& & &\checkmark& & 23.21\\
  MFM& & & &\checkmark & 22.32\\
  MFM&\checkmark&\checkmark&\checkmark&\checkmark & 22.38\\
  \hline
  MFM (with $\bm\beta :=\bm 0$)&&&&\checkmark& 22.86 \\
  MFM (with $\bm\gamma:=\bm 1$)&&&&\checkmark& 23.24\\
  \bottomrule
  \end{tabular}
\label{table_lenet_ablation}
\vspace{0mm}
\end{table}
We have studied the effectiveness of modulator network implemented on different layers of classification network and the effect of scaling parameter $\bm\gamma$ and shifting parameter $\bm\beta$ on recognition performance in Section 4.2 of the main text. Here we further conduct an ablation study with a modified LeNet \cite{lecun1998gradient} (the last convolution layer with a kernel size of 3 and a global average pooling followed it) on long-tailed Fashion-MNIST \cite{xiao2017} to corroborate the aforementioned findings are architecture-agnostic. We experiment with the long-tailed Fashion-MNIST under an imbalance factor of 1000, which is created by the same strategies in Section 4.1 of the main text. Under this setting, the meta data are copied from the current training data with 10 samples per class, other than the least frequency class with 6 samples.  The experimental settings are similar as that for CIFAR, except that no data augmentation is adopted and the initial learning rate is 0.01. The results that quantify the effectiveness of modulator network implemented on different layers of LeNet are listed in Table \ref{table_lenet_ablation}. It confirms that modulating features of the last layer (excluding the last classifier layer) achieves the best performance, no matter the last layer is convolution or fully-connected layer. As explained in the main text, this is mainly because the shallow convolution layers capture simple and general visual textures, whereas deep layers pay more attention to semantic and class-specific patterns \cite{zeiler2014visualizing}.

\end{document}